\documentclass[letterpaper, 10 pt, conference]{ieeeconf}  

\IEEEoverridecommandlockouts                            

\usepackage{cite}
\usepackage{amsmath,amssymb,amsfonts}
\usepackage{graphicx}
\usepackage{bm}
\usepackage{textcomp}
\usepackage[bottom]{footmisc}
\usepackage{xcolor}
\usepackage{hyperref}
\usepackage{subcaption}
\def\BibTeX{{\rm B\kern-.05em{\sc i\kern-.025em b}\kern-.08em
    T\kern-.1667em\lower.7ex\hbox{E}\kern-.125emX}}
    
\DeclareMathOperator*{\argmax}{arg\,max}

\usepackage{algorithm}
\usepackage{algpseudocode}
    
\newcommand{\xtraj}{\bm{x}}
\newcommand{\dxtraj}{\Delta\bm{x}}

\newcommand{\xgoal}{x_\text{goal}} 

\newcommand{\utraj}{\bm{u}}
\newcommand{\dutraj}{\Delta\bm{u}}

\newcommand{\costj}{c^{(j)}} 

\newcommand{\minimize}{\text{minimize}} 
\newcommand{\st}{\text{s.t.}} 
\newcommand{\norm}[1]{{\left|\left|{#1}\right|\right|}} 

\makeatletter
\newcommand{\pushright}[1]{\ifmeasuring@#1\else\omit\hfill$\displaystyle#1$\fi\ignorespaces}
\newcommand{\pushleft}[1]{\ifmeasuring@#1\else\omit$\displaystyle#1$\hfill\fi\ignorespaces}
\makeatother
\begin{document}
\pagestyle{plain}
\pagenumbering{arabic} 
\title{\bf Learning Deep SDF Maps Online for Robot Navigation and Exploration\\
}
\author{Gadiel Sznaier Camps$^{1}$, Robert Dyro$^{1}$, Marco Pavone$^{1}$ and Mac Schwager$^{1}$\\\\
\thanks{Toyota Research Institute provided funds to support this work. We are grateful for this support.}
\thanks{$^{1}$Department of Aeronautics and Astronautics, Stanford University, Stanford, CA 94305, USA, \{gsznaier, rdyro, pavone, schwager\}@stanford.edu }
}



\maketitle
\begingroup\renewcommand\thefootnote{\IEEEauthorrefmark{1}}
\endgroup

\begin{abstract}
We propose an algorithm to (i) learn online a deep signed distance function (SDF) with a LiDAR-equipped robot to represent the 3D environment geometry, and (ii) plan collision-free trajectories given this deep learned map. Our algorithm takes a stream of incoming LiDAR scans and continually optimizes a neural network to represent the SDF of the environment around its current vicinity. When the SDF network quality saturates, we cache a copy of the network, along with a learned confidence metric, and initialize a new SDF network to continue mapping new regions of the environment. We then concatenate all the cached local SDFs through a confidence-weighted scheme to give a global SDF for planning. For planning, we make use of a sequential convex model predictive control (MPC) algorithm. The MPC planner optimizes a dynamically feasible trajectory for the robot while enforcing no collisions with obstacles mapped in the global SDF. We show that our online mapping algorithm produces higher-quality maps than existing methods for online SDF training. In the WeBots simulator, we further showcase the combined mapper and planner running online---navigating autonomously and without collisions in an unknown environment.
\end{abstract}


\section{Introduction}

\begin{figure}[h]
\centering
\vspace{.5em}
\includegraphics[width=.9\linewidth]{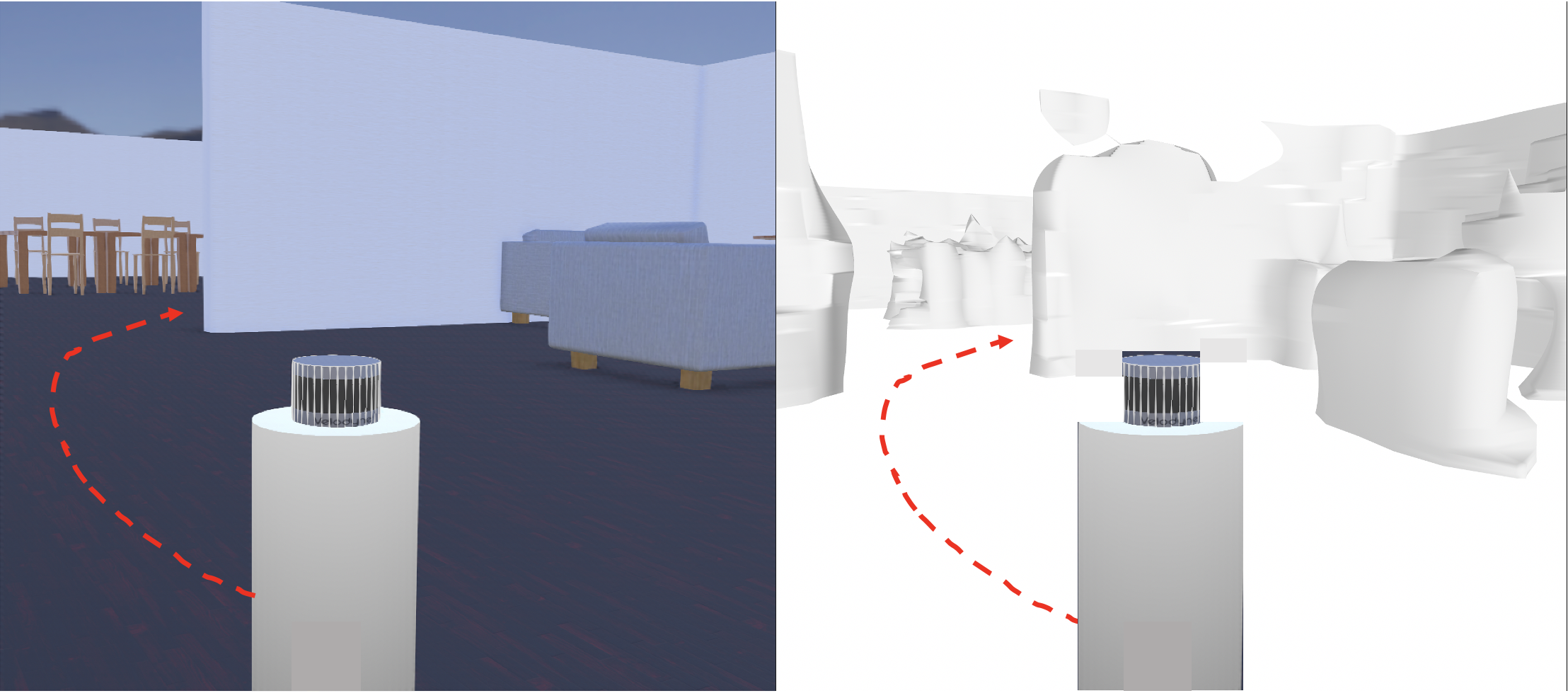}
\caption{\small A robot in the Webots simulator (right) uses our method to plan a trajectory on a map reconstructed from a deep SDF (left) learned online from LiDAR data.}
\label{fig:cover_image}
\end{figure}

We present a method to map an unknown environment by training a deep network to learn an implicit map representation online with streaming LiDAR data from a robot. We also introduce an optimization-based planner to navigate the robot through the environment given the deep network map representation as shown in Fig~\ref{fig:cover_image}.  The deep network represents the map as a signed distance function (SDF). 
In order to accommodate arbitrarily large 3D maps, we learn a sequence of local SDF maps, together with a confidence metric for each map.  We then compute the global map by fusing the local maps based on their learned confidence, thus allowing the global map to grow arbitrarily large without sacrificing map quality.  As the robot builds the map, it simultaneously uses the map to plan its trajectories.  Our trajectory planner uses sequential convex optimization to plan dynamically feasible trajectories while avoiding collisions with obstacles.  The algorithm is unique in that it accommodates a collision constraint function (i.e., the SDF map) represented as a deep network.  We use autodifferentiation to find Jacobians of the deep SDF map for use in the sequential convex optimizer.  We combine the deep SDF mapper with the sequential convex planner in a receding horizon loop to give a full robot navigation framework.

Previous approaches for building 3D maps online traditionally use voxel occupancy grids, octrees, or point clouds \cite{oleynikova2016continuous,oleynikova2017voxblox,saulnier2020information,gong2021probabilistic}, which are all costly to store in memory, suffer from inherently finite discretization, and are inconvenient for interfacing with robot trajectory planners.  Conversely, 3D scene representations from graphics and computer vision typically use triangular meshes or signed distance functions, which can efficiently represent detailed 3D geometry, but are not suited to learning online from robot sensing data.  It is only recently that researchers in machine learning and computer vision have proposed deep neural networks to learn functions that implicitly represent the environment geometry, including deep signed distance functions  \cite{park2019deepsdf,sitzmann2020metasdf,jiang2020sdfdiff},  continuous occupancy maps \cite{mescheder2019occupancy}, and NeRFs \cite{mildenhall2020nerf}.  None of these  new methods consider sensing data gathered from a robot, online network training, or robot trajectory planning using the learned map.  Further, these methods require dense data sampled over the whole environment, and use offline training that takes hours or days to obtain the desired high fidelity representation. Meanwhile, the few approaches that can learn 3D representations online do so by leveraging surface normal data \cite{yan2021continual}, which is typically unavailable in robot mapping tasks, or use RGBD images, with spatially dense, rich color and depth information \cite{sucar2021imap}. 

In contrast, comparatively little work has been done to learn SDFs online using LiDAR point clouds, which do not have color information or surface normals, and are far less spatially dense than depth camera images. Furthermore, no existing work combines online deep map learning with a trajectory planner in a receding horizon loop to give a fully autonomous deep mapping and robot navigation system. Such a system would be particularly useful in search and rescue operations after a natural disaster, during planetary exploration, exploring and mapping cave networks, remote wilderness environments, or any other scenario in which a robot needs to autonomously build and navigate within a high fidelity 3D map.  In this paper, we propose such a mapping and navigation system.  We learn a 3D map from LiDAR data incrementally without forgetting by concatenating a set of local learned SDF maps.  Simultaneously, we navigate the robot to avoid collisions using a trajectory planner that takes the deep learned SDF map as a collision constraint function.  

\section{Related Work}
\label{Sec:RelatedWork}
The proposed approach resides at the intersection between learning implicit shape representations, SDFs for robotic mapping, and robot trajectory planning. 

\emph{Learned Implicit Shape Representations:} A new trend in machine learning and computer vision is to learn a 3D shape representation implicitly as a deep neural network function.  Works that exemplify this trend include deep SDFs \cite{park2019deepsdf,atzmon2020sal,xu2019disn,sitzmann2020metasdf,jiang2020sdfdiff}, learned continuous occupancy functions \cite{mescheder2019occupancy}, and NeRFs \cite{mildenhall2020nerf,sucar2021imap}.  While these methods can represent the surface of an object with arbitrary precision, they use large volumes of dense data and require hours or days to train offline. They are not directly suitable for online training from LiDAR data, and therefore are also not useful for robot navigation.  One work, Sucar et al. \cite{sucar2021imap}, gives an algorithm for online NeRF training, but not applied to robot navigation.  NeRFs are a fundamentally different map representation than SDFs, requiring RGB or RGBD images, not LiDAR scans. Hence we do not compare to this method in our results. Common lesson from all of these deep architectures is to use an initial layer of Fourier features in the network \cite{sitzmann2020implicit}, which gives higher quality detail with fewer network parameters.  Our method builds on the seminal deepSDF work \cite{park2019deepsdf} and includes a Fourier feature layer, however our use of LiDAR data collected online, and online training for robot navigation, is unique among implicit shape learning methods.

\emph{Signed Distance Functions for Robotic Mapping:}
Mapping an environment using SDFs is a well established practice \cite{reijgwart2019voxgraph,millane2019free,daun2019large,vespa2018efficient}. However,  existing methods use discretized SDF representations, or SDF shape primitives. Furthermore, it is only recently that SDF maps represented as deep networks (deep SDFs) have begun to be applied to robotics problems. In \cite{yan2021continual}, the authors present an online learning method in an effort to bridge the gap between implicit scene representation and data collection in robotics, but do not apply their approach in the context of robotic navigation and require ground truth surface normals for training. This work recognizes the core problem of catastrophic forgetting, where old map regions are forgotten as the model's weights are fine tuned to represent new map regions, and handles the issue by exploiting experience replay by maintaining a fixed sized buffer of training data. We compare to this method in Sec.~\ref{Sec:Simulations} and find that it can still lead to catastrophic forgetting.  

Similarly, \cite{Ortiz:etal:iSDF2022} also propose an online continual learning approach, inspired by iMAP \cite{sucar2021imap}, which samples from a growing sparse set of keyframe images while learning the scene. The authors take their approach one step further by evaluating the usefulness of their approach for downstream navigation and manipulation tasks by examining the collision costs of a chosen trajectory as the scene is learned. However, they also do not integrate their approach with a trajectory planning algorithm.

We propose a different method to combat catastrophic forgetting inspired by patch-based offline SDF training \cite{tretschk2020patchnets, chabra2020deep}.  We save a growing set of trained local SDF maps and continually train a new local SDF map from a training dataset that is updated as the robot moves.  This yields training rates that remain constant over time, but allows for a growing global map of arbitrarily high quality, since the cached local maps can be fused to yield a global map.

\emph{Trajectory Planning:} Trajectory planning for a robot given a signed distance function map has received considerable attention in the literature \cite{oleynikova2016continuous,oleynikova2017voxblox,saulnier2020information,gong2021probabilistic}. The two main types of approaches rely on (i) gradient based trajectory optimization, like CHOMP \cite{oleynikova2017voxblox,oleynikova2016continuous}, and (ii) trajectory sampling, using a random sampling or sampling with receding horizon pruning, until the approximately optimal trajectory is obtained \cite{saulnier2020information, gong2021probabilistic}. However, these algorithms do not use a deep learned SDF, and rarely involve actively exploring the environment to populate the map \cite{oleynikova2017voxblox, saulnier2020information}.

Since we focus on a learned signed distance representation, but require local optimality of the generated trajectory, we take an approach most closely related to gradient-based model-based receding horizon planning for nonlinear dynamics and a cost function, explored in \cite{betts1998survey, camacho2013model, bonalli2019gusto}. We make use of the implementation in \cite{dyro2021particle} which relies on sequential convex programming (SCP) \cite{quoc2012real} to handle nonlinearities. Our approach introduces a novel use case in that the deep network represented SDF requires the underlying SCP optimization algorithm to effectively use Jacobians for the deep network obtained via automatic differentiation.

\section{Problem Set up}
\label{Sec:ProblemSetUp}
We consider a robot with pose $x_t\in\text{SE}(N)$ at a discrete time instance $t$, where $\text{SE}(N)$ is the special Euclidean group of rigid body transformations in $N$-dimensional Euclidean space.  Our framework applies both to $N = 2$ or $N = 3$.  The robot has a body $\Omega_{\text{robot}} \subset \mathbb{R}^3$, consisting of the set of points in 3D space that are occupied by the robot.  We denote the dependence of the robot's body on its pose as a set-valued function $\Omega_{\text{robot}}(x_t)$.  The robot body is embedded in an ambient 3D space together with obstacles. Collectively, we call these obstacles the ``map,'' and we denote the set of all points occupied by these obstacles as $\Omega_{\text{map}} \subset{\mathbb{R}^3}$.  We denote the surface (i.e., the boundary) of the map as $\partial\Omega_{\text{map}}$.  We assume that both the robot body and the map are closed sets, and that the robot knows its own pose, but it does not know the map. However, the robot is equipped with a LiDAR sensor with which it can sample points on the surface of the map. The goal of the robot is to navigate through the map without collisions (so $\Omega_\text{robot}(x_t) \cap \Omega_\text{map} = \emptyset$ for all $t$), with a trajectory that minimizes a cost function, which may include per stage, control  and final costs.

For simplicity, we assume that the robot consists of a single rigid body, and that it moves according to the kinematic equation $x_{t+1} = f(x_t, u_t)$, with commanded input vector $u_t \in \mathbb{R}^m$.  This model is approximately valid for, e.g., a differential drive robot or quadrotor UAV.  However, the tools we develop could be extended to multi-body robots with dynamic state and more complex or under actuated dynamics, such as humanoid robots or multi-link manipulators.

\subsection{Signed Distance Function} The core idea of this paper is to model the unknown map $\Omega_\text{map}$ as a signed distance function (SDF), learned  from LiDAR data. The SDF for a set $\Omega$ is a function, $S(p)$, that when given a spatial point $p$ in $\mathbb{R}^3$, returns the corresponding distance to the closest location on the boundary of $\Omega$. The sign of the distance then specifies whether $p$ is inside (negative) or outside (positive) the set.  Specifically, we have $S(p) = +d(p,\partial\Omega) $ if $p \notin \Omega$ or $S(p) = -d(p,\partial\Omega)$ if $p \in \Omega$
where $\partial\Omega$ is the boundary of the set, and the function $d(p,\partial \Omega) = \min_{\{q\in \partial\Omega\}}\|p - q\|_2$ gives the Euclidean distance from the point $p$ to the nearest point in  $\partial \Omega$. 
If the original set $\Omega$ is closed, then it is conveniently retrieved as the zero sub-level set of the SDF, $\Omega = \{p \mid S(p)\le 0\}$, otherwise this gives the closure of the original set.  If the boundary of the set is piece-wise smooth (which is often the case in robotic mapping), the SDF is differentiable almost everywhere.  Wherever the gradient of the SDF exists, it satisfies the Eikonal equation $||\nabla S|| = 1$ \cite{sitzmann2020implicit}.  In addition, the maximum of multiple SDFs gives the SDF of the intersection of their underlying sets, and the minimum gives the SDF of their union, making it straightforward to construct complex representations from simple signed distance functions, or to check for collisions between multiple objects with SDF representations \cite{benton2019furthergraphics}.  We write the SDF for the map as $S_\text{map}(p)$, and for the robot at $S_\text{robot}(p, x_t)$.  Note that the robot's SDF depends on both the robot's pose $x_t$, and the query point in space $p$.  We represent the learned SDF as a neural network, which we write $\hat{S}_\text{map}(p, \theta)$, where $\theta$ is the vector of neural network parameters.  We assume the robot knows the SDF for its own body $S_\text{robot}(p, x_t)$.



\section{SDF Online Learning }
\label{Sec:OnLineLearning}

Two challenges in training a neural network to represent an SDF online are (i) that the effective complexity of the environment is unknown \emph{a priori} and so, the required size of the network, and (ii) a large network is too slow to train online. To overcome these issues, we train a sequence of small local SDFs, $\hat{S}_i(p, \theta_i)$, $i = 1, 2, \ldots$, together with a spatial confidence metric, $C_i(p, \theta_i)$, $i = 1, 2, \ldots$. When one of these local SDFs reaches its learning capacity, it is cached, and a new local SDF is initialized using the previous local SDF's weights.
The global SDF is defined as the the value of a local SDF with the highest confidence at the query point, $\hat{S}_\text{map}(p) = \hat{S}_{i^*}(p, \theta_{i^*})$, where $i^* = \argmax_i\{C_i(p,\theta_i)\}$. 

\subsection{Model Architecture}
\begin{figure}[h]
\centering
\vspace{-1em}
\includegraphics[width=.49\linewidth]{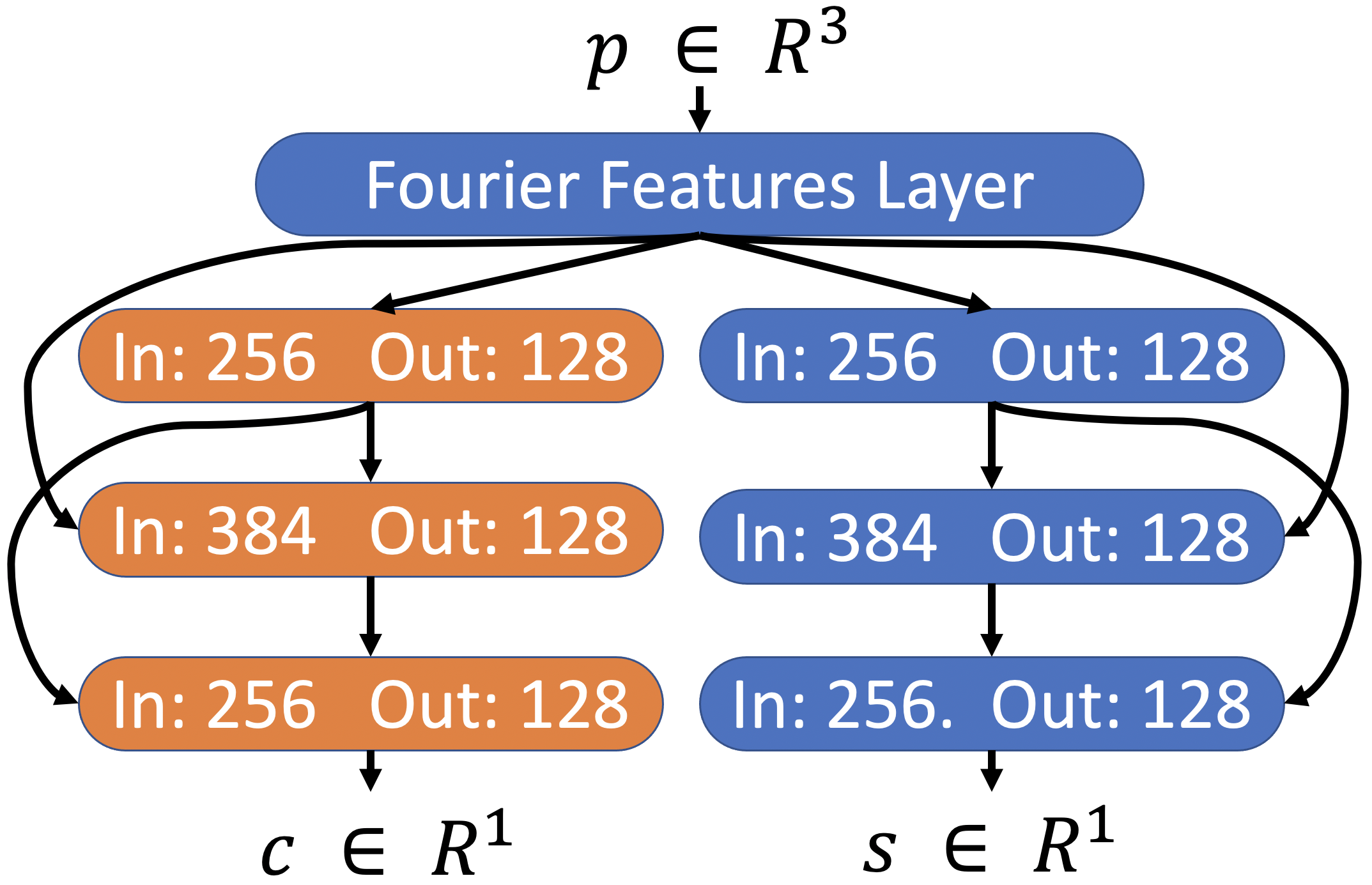}
\includegraphics[width=.49\linewidth]{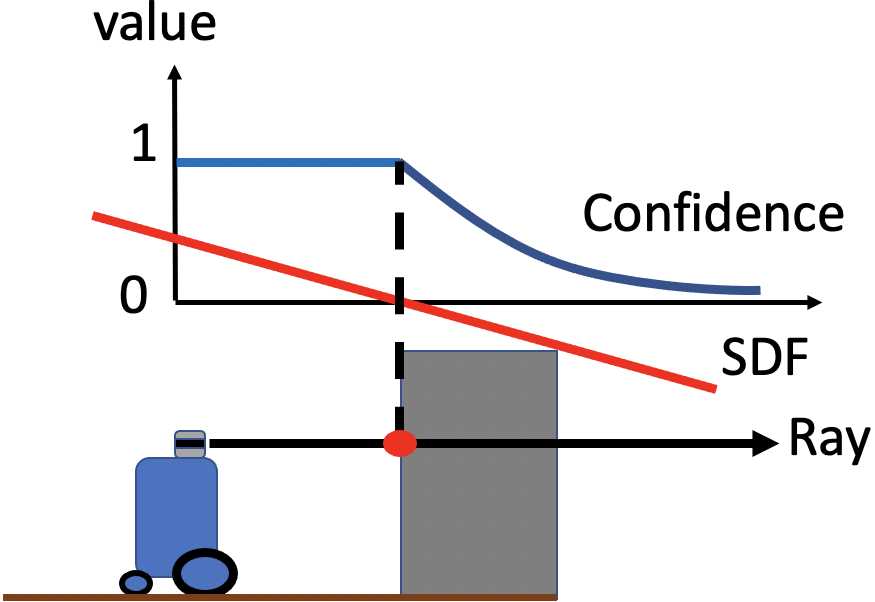}
\caption{\small \textbf{(Left)} Model architecture. It takes as input a point p and outputs the scalar values $s$ and $c$, which are the signed distance and the corresponding confidence respectively. The first layer is the Fourier feature layer that encodes the input into frequency dimensions. \textbf{(Right)} In free space the SDF is known with high confidence from a single LiDAR ray. After penetrating the obstacle, we label sample points along the ray with exponentially decreasing confidence, representing our uncertainty about object depth. }
\label{fig:confidence_diag}
\end{figure}

We learn the SDF and the confidence metric in a combined neural network architecture shown in Fig.~\ref{fig:confidence_diag}.
The local SDF and its confidence metric are each represented using a small 3-layer fully connected neural networks with residual skip connections. The confidence model uses tanh as its activation function while the SDF model use Tanhshrink as its activation function. The two networks share a single Fourier feature encoding layer.  We, as well as many others \cite{tancik2020fourier, sucar2021imap, sitzmann2020implicit}, have observed that encoding the spatial input using a Fourier feature layer improves a network's ability to learn fine details.  The encoding layer takes a spatial input point in $R^{3}$ and returns a frequency vector output in $R^{128}$.  The networks then output two values, $\hat{S}_i$ and $\hat{C}_i$ which correspond to the signed distance and the confidence for a given $p$.


\vspace{-1em}

\subsection{Data Pre-Processing}
It is necessary to augment LiDAR scan data to learn an SDF, since the LiDAR measurement only sees the obstacles' surface---where the SDF is zero. Typically, the data is augmented with known positive distance points in free space, and corresponding negative distance points inside the obstacle. Other works using deep networks to represent SDFs, e.g. \cite{park2019deepsdf,sitzmann2020metasdf}, propose various data augmentation schemes. We outline our own augmentation procedure as follows.


We uniformly sample points along each LiDAR ray and produce approximate depth and confidence labels for all points after the surface of an object. Points in free space are labeled with SDF values obtained by taking the minimum distance from the sample point to all existing observed surface points and are given a confidence of $1$. This gives a provable over-approximation of the true SDF value at that point, as known surface points are a subset of the true continuous set of surface points. We calculate the distance to the nearest surface point using a K-D Tree with 50 leaf nodes. A naive method to label free space SDF values is to use the distance from where its ray terminates. We find that in practice this gives badly incorrect distance labels which leads to poor learned SDF quality.

To obtain negative distance data, we sample points along the LiDAR ray beyond the surface where the ray terminates, as we expect objects to have a finite depth. We assign sample points with exponentially decreasing confidence labels, representing decreasing uncertainty about the object's depth. We approximate the depth into the object by computing the distance to the nearest surface point as done for free space points. This results in confidence labels for data such that
\begin{equation}\label{eqd_1}
    C(p) = \begin{cases}
    1,               & \text{if } S_\text{map}(p) \geq 0\\
    \frac{b^{w}-1}{b-1} + 1e^{-7} & \text{otherwise}, 
\end{cases}
\end{equation}
where $w = 1 - d(p)/d_{max}$ is a normalized distance between 0 and 1, $d(p)$ is the distance away from the surface along the ray, $d_{max}$ is a maximum distance along the ray, and $b$ is a hyperparameter.  This confidence labeling scheme is shown in the right of Fig.~\ref{fig:confidence_diag}


\subsection{Data Ingestion}
During training, a fixed sized synthetic and surface dataset is maintained for training the network. At each time step $t$, the model receives a new surface LiDAR scan which is accepted or rejected based on its similarity with currently stored data, as determined by the directed Hausdorff distance. If accepted, the algorithm ingests the data through the following steps: (1) The new scan is separated into obstacle and floor points, and the floor points discarded. (2) The remaining scan points are used to generate synthetic data by shooting rays from the current robot origin to each surface point and sampling along them as described earlier. (3) The signed distance labels are calculated by finding the distance to the nearest surface point for each  sample. (4) The associated confidence labels are generated through the weighting process described above. (5) The new and old surface data are concatenated and sub-sampled by uniformly choosing points at random, until the resulting dataset is of the desired size. 



\subsection{Local SDF Training}
To construct a global map of the environment, several local maps are trained using the architecture described above. At each iteration, a local SDF is trained using a set of labeled data points $(p,s,c)$, where $p$ is the spatial sample location, $s$ is the SDF label, and $c$ is the confidence label.  At each data point we use the loss function
\begin{align}
    L(p,s,c, \theta_i) = 10^{4}L_{H_s}(p,s, \theta_i) + 10^{4}L_{H_c}(p,c, \theta_i)\\ + L_E(p,s, \theta_i) + 10^{-3}L_R(\theta_i),
\end{align}
and sum over all data points, where $L_{H_s}$ and $L_{H_s}$ are the Huber loss \cite{paszke2019pytorch} for the signed distance and confidence values, respectively.  Specifically, for the signed distance this loss is defined as
\begin{equation} \label{eqmc_2}
    L_{H_s}(p,s,\theta_i) = \begin{cases}
    0.5(\hat{S}_i(p, \theta_i) - s)^2,  \quad \text{if } |\hat{S}_i(p, \theta_i) - s| < \delta\\
    \delta(|\hat{S}_i(p, \theta_i) - s| - 0.5\delta), \quad \text{otherwise}, 
\end{cases}
\end{equation}
where $\hat{S}_i(p, \theta_i)$ is the predicted signed distance for the $i$th local SDF at sample point $p$.  The Huber loss for the confidence is defined analogously.  $L_E = (||\nabla_{p} \hat{S}_i(p,\theta_i)||-1)^2$ is the Eikonal loss to enforce a unit norm gradient on the SDF, and $L_R = ||\theta_i||_{\infty}$ (where $\theta_i$ is the vector of network weights) regularizes the Lipschitz constant of the network to promote smoothness. Once the network has converged to a loss within a threshold, the model is then saved and the training of a new map is initiated. To speed up convergence, the new map uses the previous model weights as a starting point. At the same time, to prevent the new map from being too similar to the previous model, a waiting period is initiated where training the model is paused until the directed Hausdorff distance between a new LiDAR scan and the previous local map is greater than some threshold. In this way, the number of redundant local maps is reduced.  The global map is constructed at any time by finding the most confident  trained local map at a given query point, $\hat{S}_\text{map}(p) = \hat{S}_{i^*}(p)$, where $i^* = \argmax_i \{C_i(p)\}$.

\section{Trajectory Optimization with the Deep SDF}
\label{Sec:TrajectoryPlanning}
We propose a trajectory optimization approach to have the robot autonomously navigate the unknown environment with the learned map. We assume known robot dynamics, and optimize over control actions that give an exploratory trajectory that (i) avoids collisions with the learned representation of the environment and (ii) navigates towards unexplored regions. Unlike feedback control, trajectory optimization allows us to reason about whole robot trajectories and so explicitly plan around obstacles, not just avoid them.  As is typical for receding horizon control, after planning, we execute the first action and replan at the next control time step. 
\begin{equation}
\label{eqn:fhoptcontrol}
\begin{split}
\minimize_{\utraj}~~& c \big(\xtraj, \utraj\big) \\
\st ~~ & x_0 = x(0) \\
& x_{t+1} = f(x_t,u_t) \\
 & u_t \in \mathbb{U} ~~~~ \forall t
\end{split}
\end{equation}
where $\xtraj = (x_0, \dots, x_{T+1})$ and $\utraj = (u_0, \dots, u_T)$ is the state and control trajectory. $\mathbb{U}$ denotes the feasible set of controls at time $t$, which we take to be a box constraint $u_\text{lower} \leq u_t \leq u_\text{upper}$. We incorporate a collision penalty directly within $c(\cdot)$ using exact penalty functions \cite{han1979exact} to ensure persistent feasibility of the control problem.  This formulation gives exact collision avoidance when feasible, but allows for the smallest possible penetration when a collision is impossible to avoid. The majority of the behavior specification for the robot controller is contained in the cost function definition.

\subsection{Cost Specification}
Taking the trajectory optimization approach allows us to specify the desired behavior of the robot using the cost function $\costj \big( \xtraj, \utraj \big)$. We set out to capture the desired behavior by including following specifications: (1) exploratory goal state reaching, (2) control energy minimizing, (3) control slew minimizing and (4) avoidance of collisions with obstacles in the learned SDF map. We take a weighted linear combination of the 4 objective terms
\[
c\big(\xtraj, \utraj\big) = 
w_1 c_{1}\big(\xtraj, \utraj\big) 
+ w_2 c_{2}\big(\xtraj, \utraj\big) 
+ w_3 c_{3}\big(\xtraj, \utraj\big) 
+ w_4 c_{4}\big(\xtraj, \utraj\big) 
\]

We formulate both (1) and (2) as a quadratic penalty around a reference. In particular 
\begin{align}
c_1\big( \xtraj, \utraj \big) &= \sum_{t=0}^{T+1} \big( x_t - \xgoal \big)^T Q \big( x_t - \xgoal \big) \\
c_2\big( \xtraj, \utraj \big) &= \sum_{t = 0}^Tu_t^T R u_t
\end{align}
where $Q$ and $R$ are positive definite matrices; in practice we select diagonal matrices with per-state/control weight. The goal state, $\xgoal$ is selected by scanning the map for the point location with high uncertainty of the learned map---online selection of the goal state as the region with high learned map uncertainty is how we guide exploration. 

Minimizing control slew corresponds to minimizing the energy of the difference in adjacent control states
\begin{equation}
c_3 \big( \xtraj, \utraj \big) = \sum_{t=1}^T \norm{u_t - u_{t-1}}_2^2
\end{equation}

Finally, we formulate the (4) collision avoidance term as an exact penalty function
\begin{equation}
\label{eqn:c4}
c_4 \big( \xtraj, \utraj \big) = \sum_{t=0}^{T+1}\max\{-d(x_t), 0\}
\end{equation}
where $d(\cdot)$ is the signed distance of the closest point on the robot to any obstacle in the environment. This distance itself is defined as 
\begin{equation}
d(x_t) = \min_{p\in \Omega_\text{robot}(x_t)} \hat{S}_\text{map}\big(p\big).
\end{equation}
In practice we pick a finite number of sample points around the robot body and make use of automatic differentiation to approximate the Jacobian $\nabla_{x_t}d(x_t)$ within the solver.\footnote{In fact, the implicit function theorem \cite{gould2016differentiating} allows for the analytical derivation of $\nabla_{x_t}d(x_t)$, which can be used to optimize for collision avoidance without requiring sample points of the robot body.  This is an interesting direction for future work.} 


\subsection{Implementation}

In order to solve the MPC problem over nonlinear constraints and the nonlinear objective function, we make use of sequential convex programming (SCP) \cite{quoc2012real}, where we (i) iteratively linearize the dynamics around the current trajectory solution, $\xtraj$, $\utraj$ and penalize the deviation from the linearization to ensure convergence and asymptotic dynamic feasibility, and (ii) linearize the non-convex elements of the cost function. This reformulation is stated in Equation \eqref{eqn:convexified_pmpc}.

\begin{equation}
\label{eqn:convexified_pmpc}
\begin{aligned}
\min_{\dxtraj, \dutraj} & \bigg( c_\text{cvx}(\mathbf{x} + \Delta\mathbf{x}, \mathbf{u} + \Delta\mathbf{u}) \\
& ~~~~~~~ + \nabla_x c_\text{ncvx} \Delta\mathbf{x} + \nabla_u c_\text{ncvx} \Delta\mathbf{u} \\
& ~~~~~~~~~~ + \rho_x\| \Delta\mathbf{x} \|_2^2 + \rho_u\|\Delta\mathbf{u}\|_2^2 \bigg) \\
\mathrm{s.t.}~~& \Delta x_{t+1} = \nabla_x f_t \Delta x_t + \nabla_u f_t \Delta u_t \quad 
\\
& u_t + \Delta u_t \in \mathbb{U} 
~~~~~~~
\forall t \in \{0,\ldots,T\} \\
\end{aligned}
\end{equation}

The primary element of the cost function requiring linearization is the learned SDF map. We use automatic differentiation to obtain those derivatives. This can be done efficiently because we only require derivatives wrt a finite number of states which requires only a single backward pass through the network per state (3 in our case). The concatenation of the learned SDF maps via the $\max$ operator makes the network locally not smooth at some points, so in practice we use the softmax reformulation with a sharpness parameter $\alpha$. We note that even for partially learned networks this smoothing and linearization approach works very well
\[ 
\hat{S}_\text{map}(p, \theta) \approx 
\left( \sum_i^n \hat{S}_i(p, \theta_i) e^{\alpha \hat{C}_i(p, \theta_i)} \right) /
\left( \sum_i^n e^{\alpha \hat{C}_i(p, \theta_i)} \right)
\]
where we set $\alpha = 10^2$. 

Both relaxations allow us to cast the MPC problem as a quadratic program (QP) for which fast solvers exists. We make use of the Open Source Quadratic Program (OSQP) solver \cite{stellato2020osqp}. Our method introduces only two hyperparameters: state and control quadratic deviation penalties, $\rho_x$, $\rho_u$, that ensure the quality of the linearization. We refer the reader to \cite{dyro2021particle} for discussion about the convergence and implementation details of our method. 
The final SCP MPC algorithm can be found in the Algorithm \ref{alg:scp_pmpc} block.

\begin{algorithm}[t]
\caption{\centering SCP MPC}
\label{alg:scp_pmpc}
\begin{algorithmic}[1]
\Require{Initial state $x^{(0)}$, dynamics $f$, trajectory guess $\xtraj, \utraj$}
\Require{Hyperparameters $\rho_x, \rho_u$}
\Require{Solution tolerance $\epsilon$}
\Repeat
\State $\bar{f}_t, \nabla_x f_t, \nabla_u f_t \gets$ Linearize dynamics at $\xtraj, \utraj$;
\State Split cost
into convex and not- parts $c_{\text{cvx}}, c_{\text{ncvx}}$\;
\State $\nabla_x c_{\text{ncvx}},\, \nabla_u c_{\text{ncvx}}
\gets$ Linearize non-cvx cost at $\xtraj, \utraj
$\;
\State $\Delta\mathbf{x}, \Delta\mathbf{u} \gets$ Solution to convex prob. \eqref{eqn:convexified_pmpc}
\State $\xtraj, \utraj \gets \xtraj + \Delta \xtraj, \utraj + \Delta \utraj$ 
\Until{$\| \Delta\mathbf{x} \| + \| \Delta\mathbf{u} \| < \epsilon$}
\State \Return $\xtraj, \utraj$
\end{algorithmic}
\end{algorithm}

\section{Simulation Results}
\label{Sec:Simulations}
     
\begin{figure*}[h]
\centering

\begin{tabular}{ccccc}
Ground Truth & Ours & Fine-tuning & Batched & Cont. Neural Map\\
\includegraphics[width=.17\linewidth]{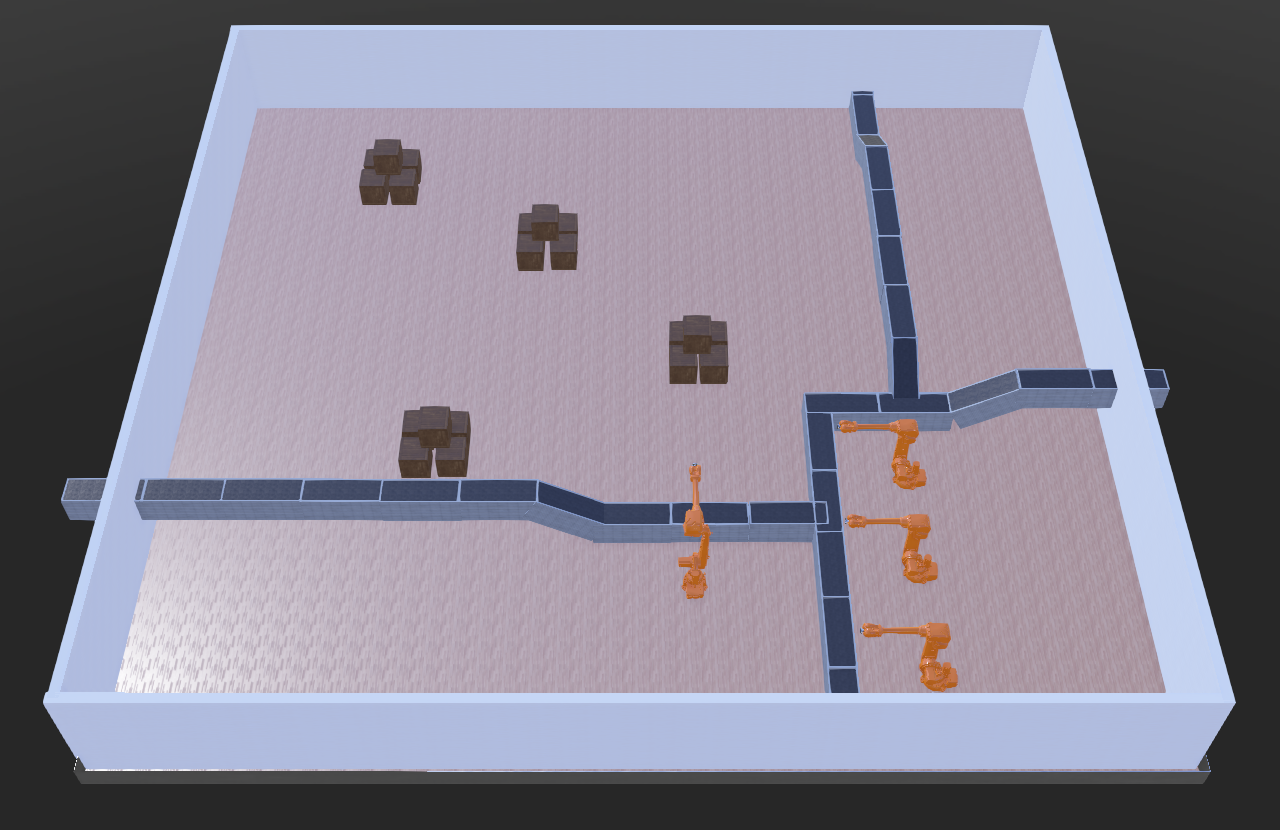}
&
\includegraphics[width=.17\linewidth]{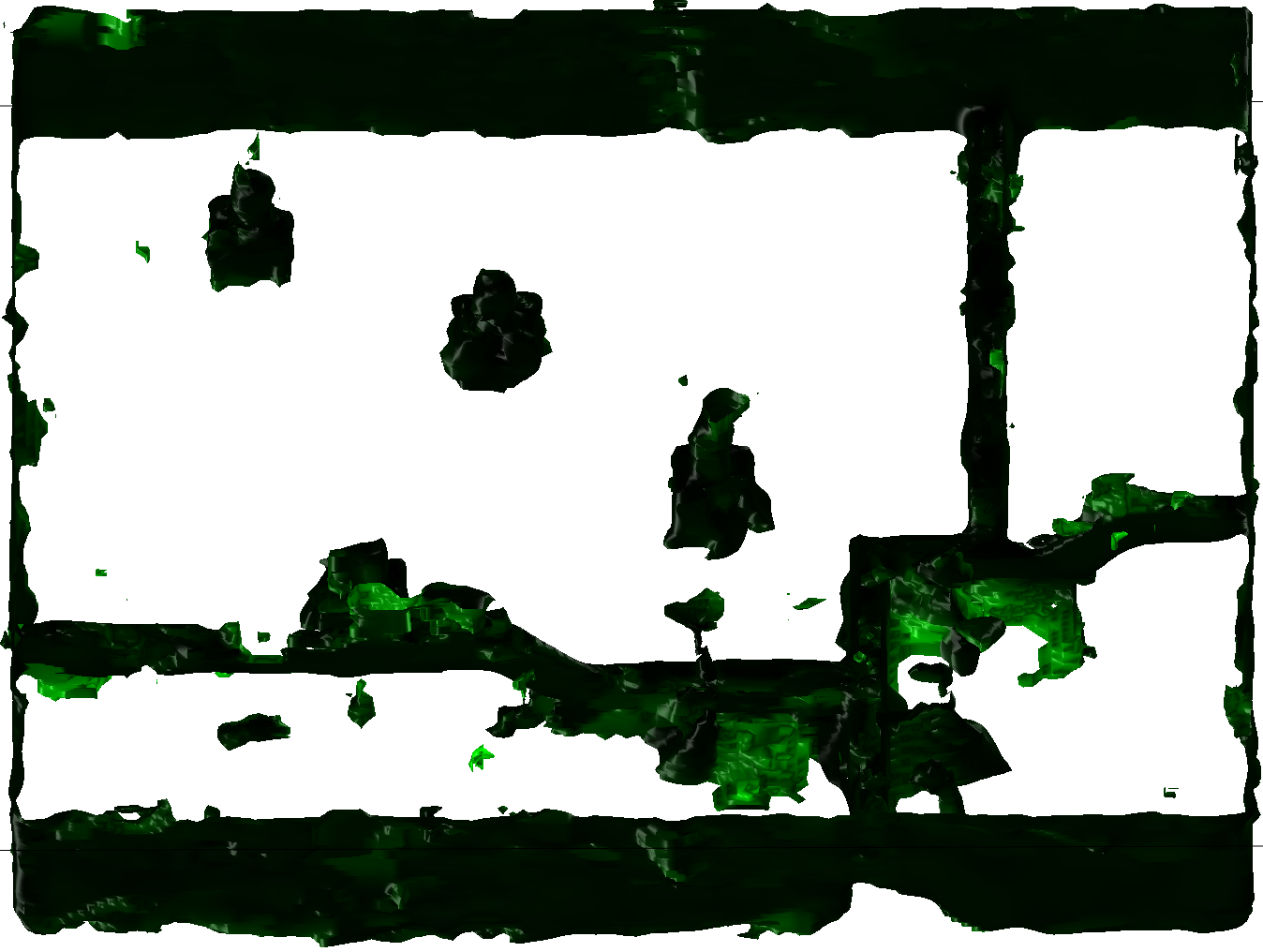}
&
\includegraphics[width=.17\linewidth]{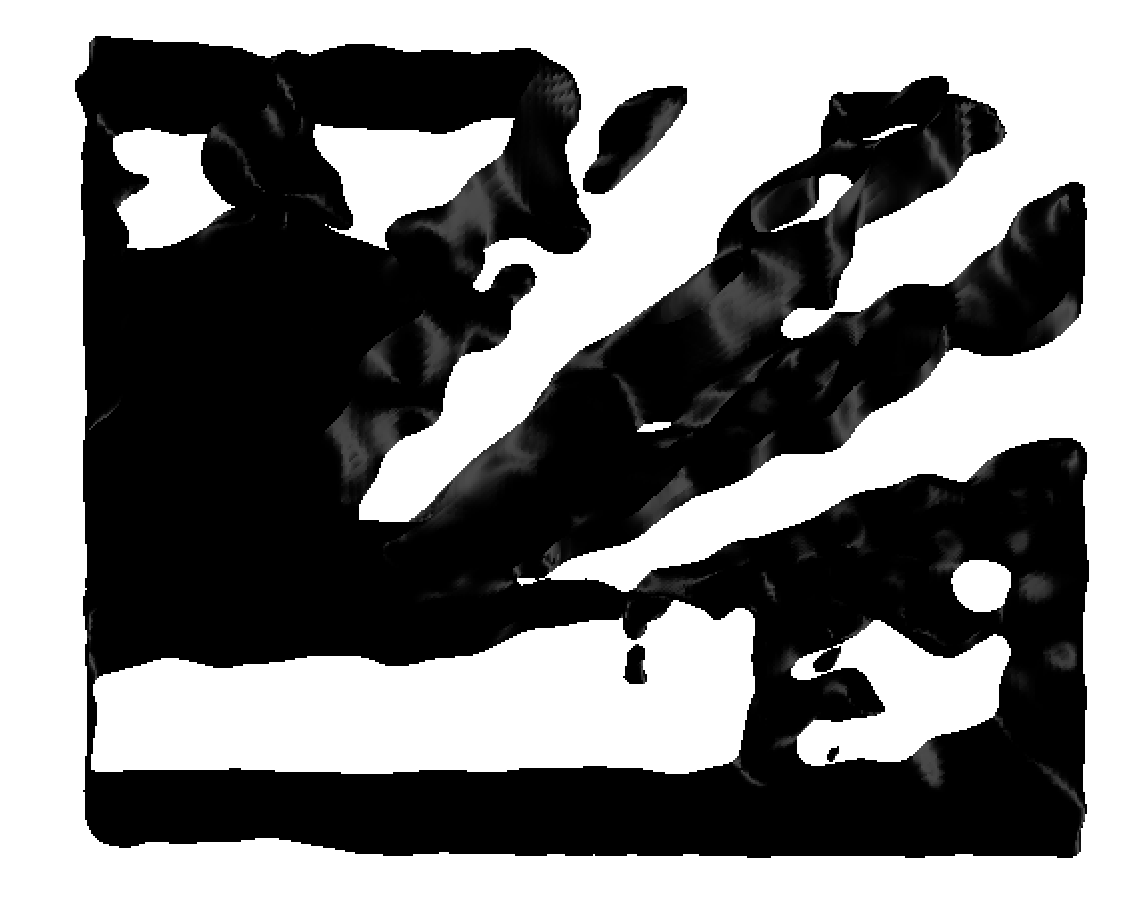}
&
\includegraphics[width=.17\linewidth]{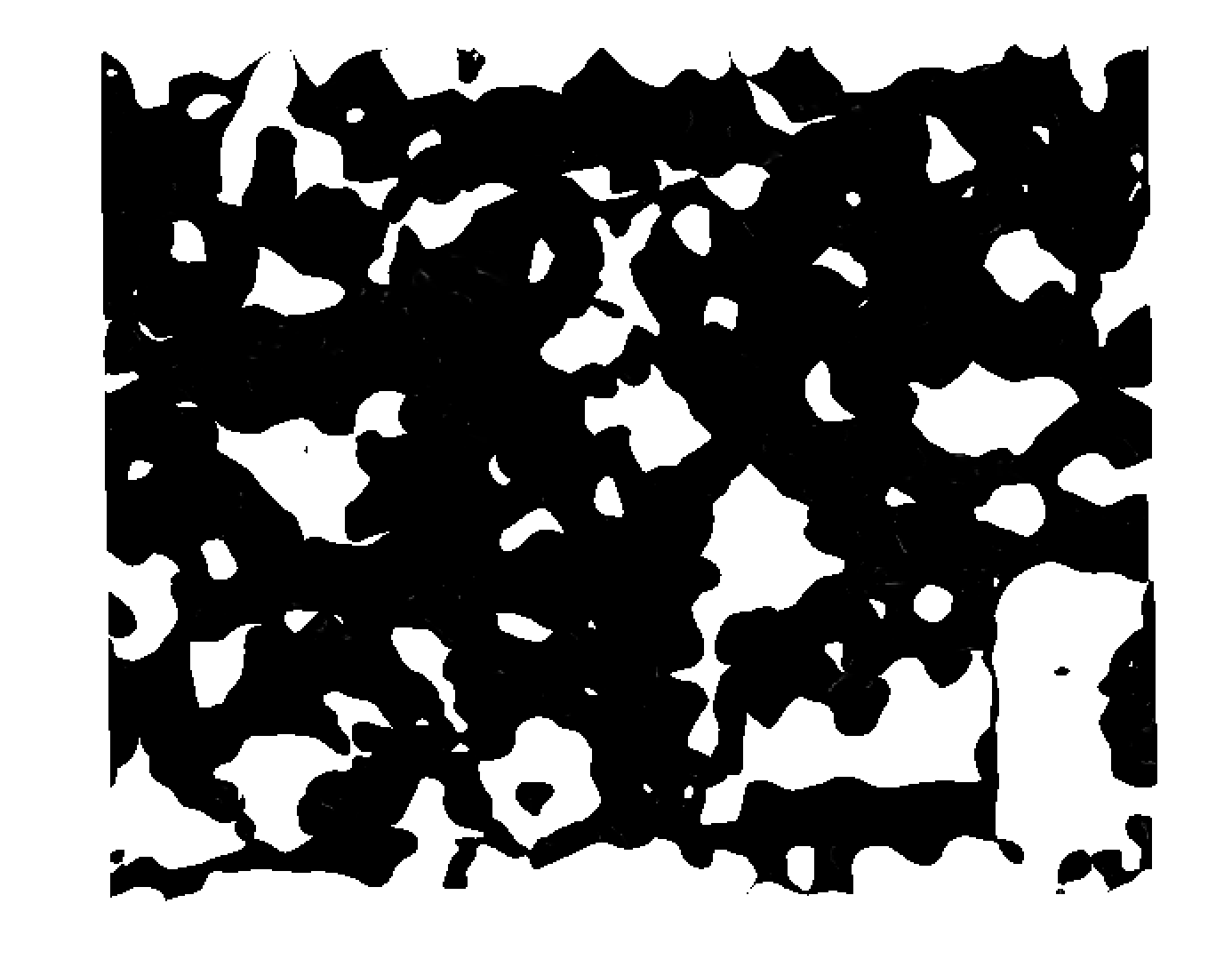}
&
\includegraphics[width=.17\linewidth]{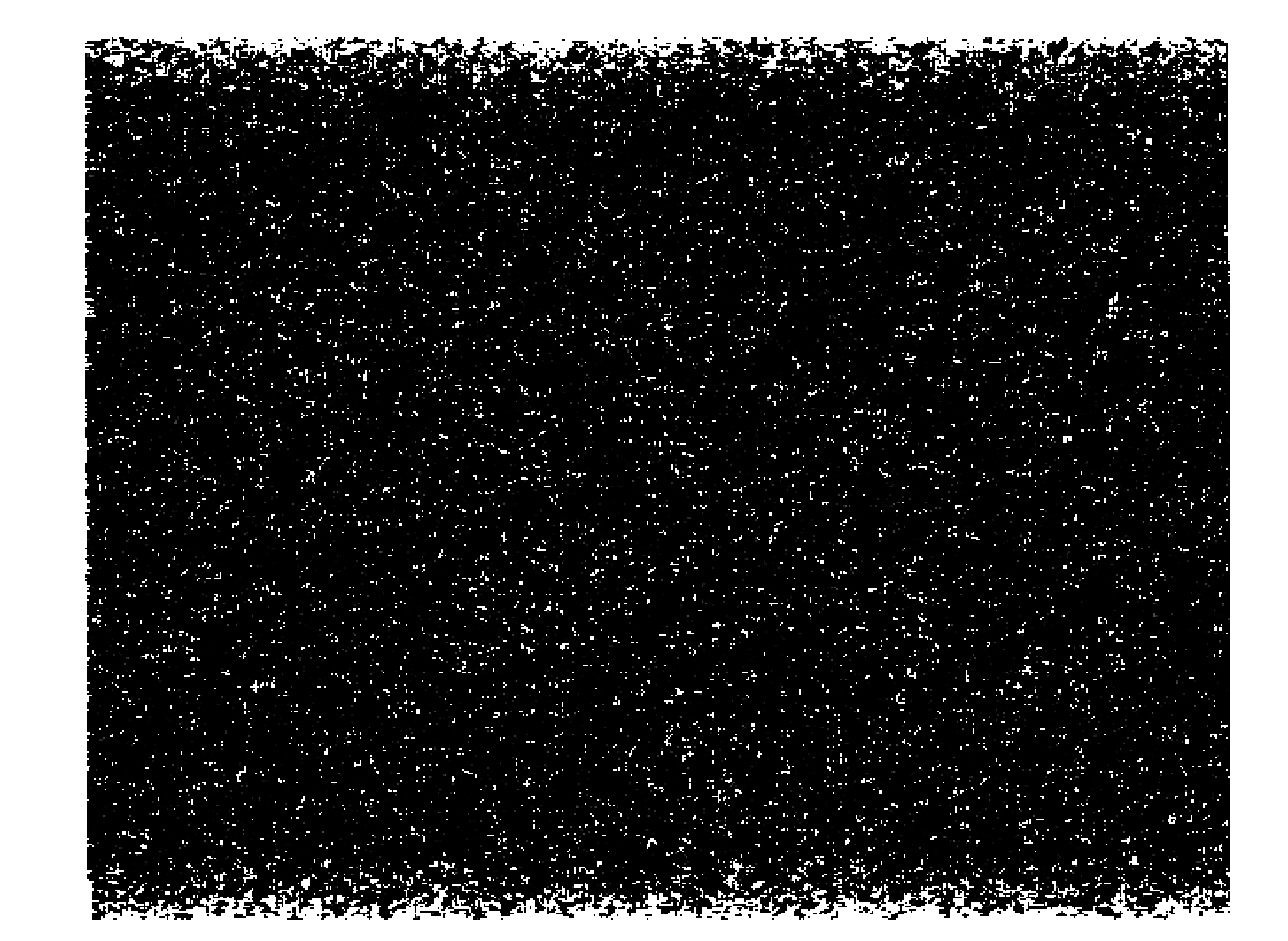}\\

\includegraphics[width=.17\linewidth]{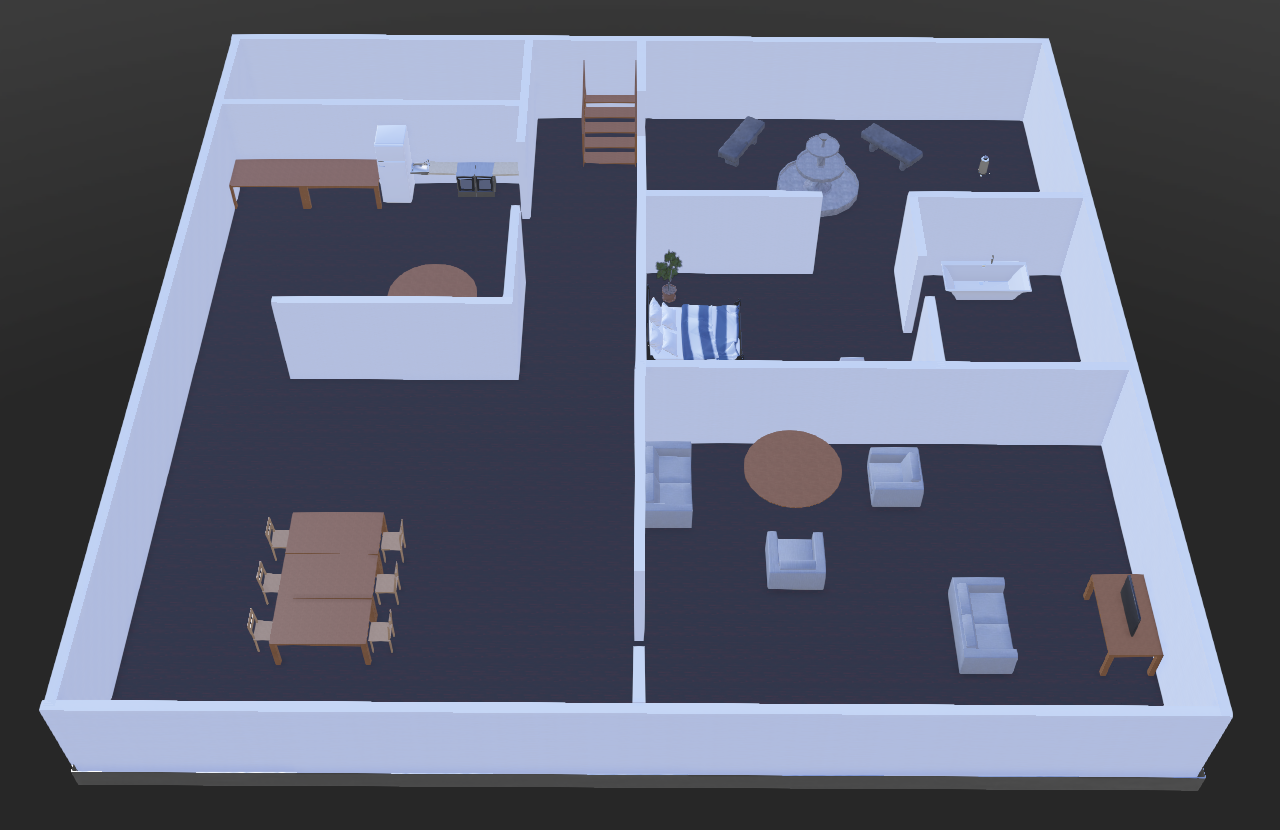}
&
\includegraphics[width=.17\linewidth]{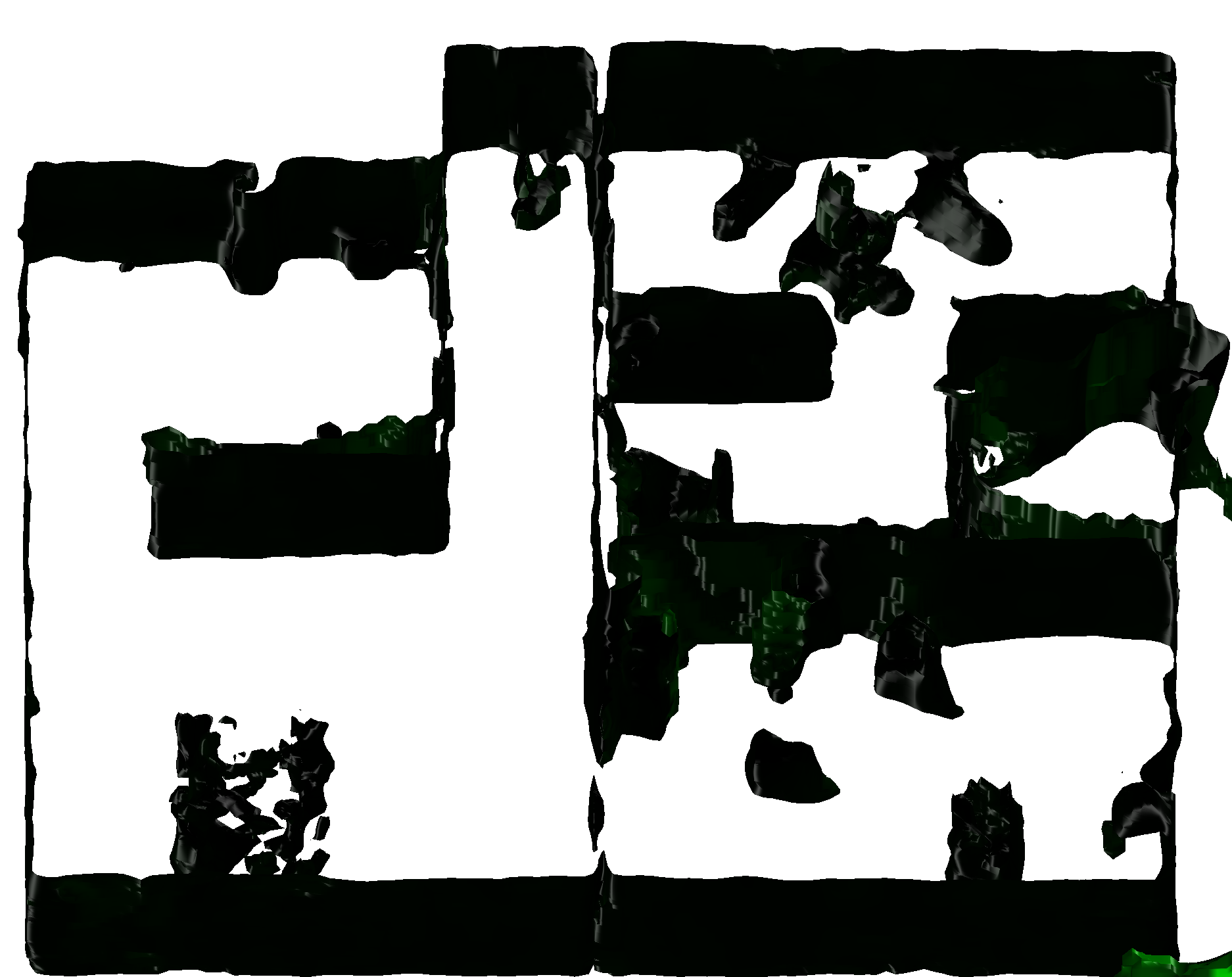}
&
\includegraphics[width=.13\linewidth]{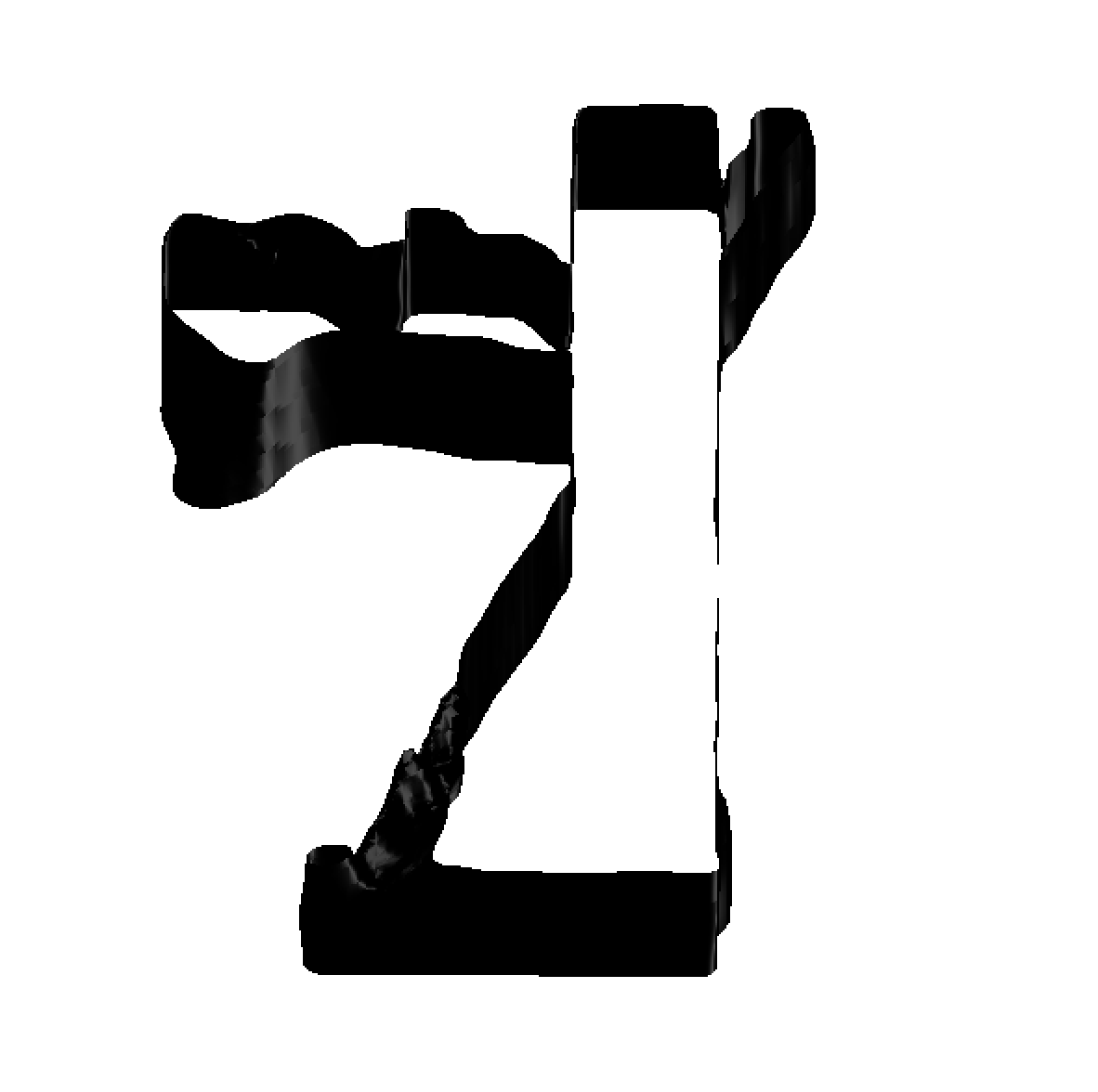}
&
\includegraphics[width=.17\linewidth]{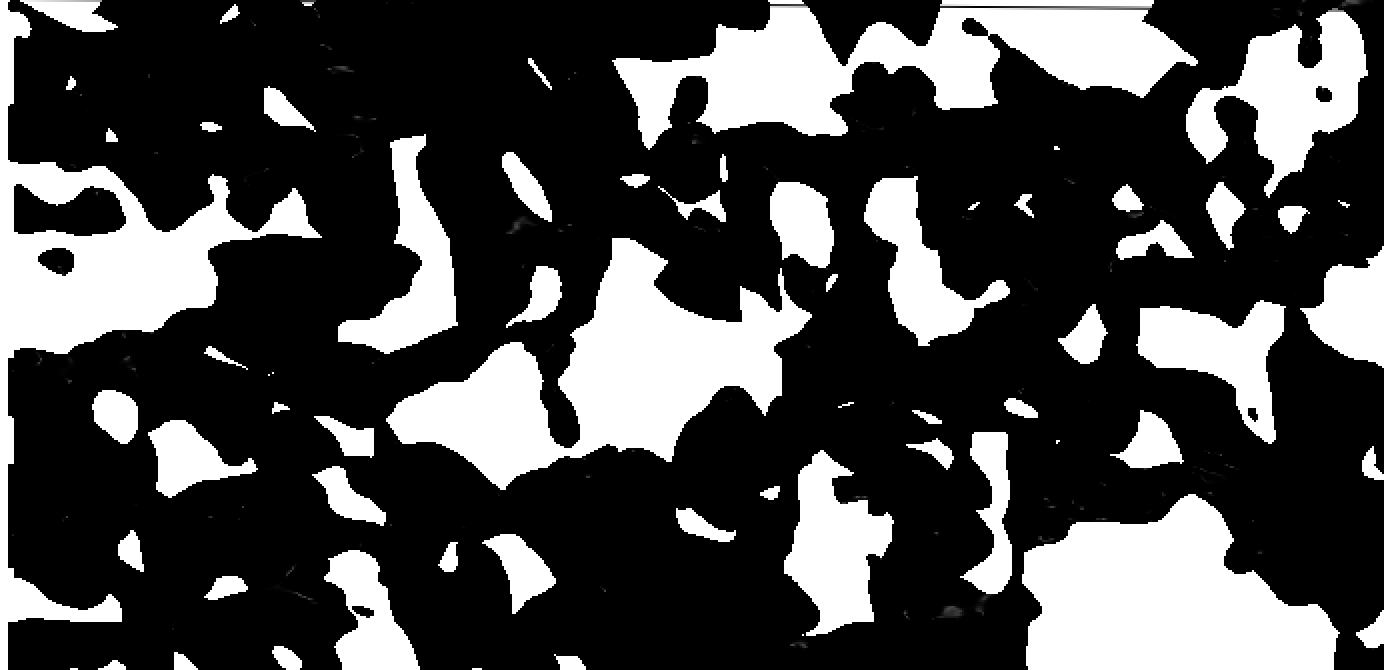}
&
\includegraphics[width=.17\linewidth]{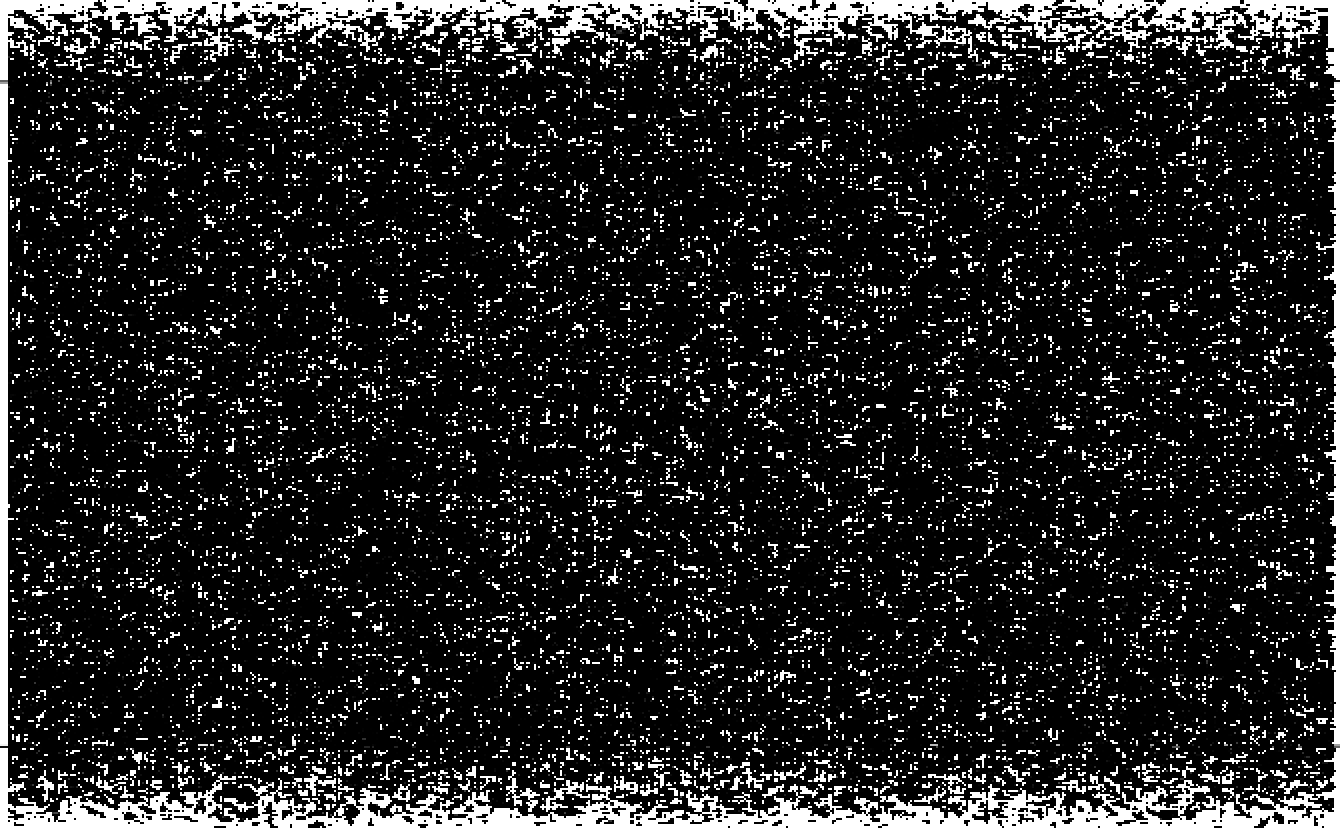}

\end{tabular}
\caption{Zero levelsets of learned SDFs of a factory (top) and a house (bottom) environment for our method, compared with three online SDF learning benchmarks from \cite{yan2021continual}.  Fine-tuning suffers from catastrophic forgetting, Batched cannot learn a map in reasonable time, and Continual Neural Mapping \cite{yan2021continual} fails to produce a meaningful map without surface normals.  Only our method produces a navigable 3D map online from LiDAR data.}
\label{fig:final_meshes}
\end{figure*}
%
%

To verify the correctness of our approach, we perform mapping experiments using the Webots simulator \cite{michel2004cyberbotics}, mapping a factory floor plan and a house floor plan. 
We select these two environments because they represent likely autonomous application scenarios and show the flexibility of our approach. In each scenario, a single robot with differential drive dynamics and a LiDAR sensor drives around the environment autonomously, using the trajectory optimizer, and gathering environment scans. We use our algorithm to construct the desired global map with LiDAR coming in at 2 Hz.
We train the model using AMSGrad with a learning rate of $10^{-2}$ and a weight decay of $10^{-4}$ and we maintain a fixed-size data set of 200k surface points and 300k augmented points. We train with a batch size of 10k mixed points and we weigh the data by inverse distance from the surface.


As in \cite{yan2021continual}, we compare our approach against three baselines: (i) fine-tuning, (ii) batch re-training, and (iii) continual neural mapping. For (i) the model weights are retrained at every time step with current LiDAR scan. For (ii) the model is trained every time step on the full data from the start to the current time step. We terminate (ii) after 20 minutes of training, because the online application demands fast training.
We base the implementation of (3) on \cite{yan2021continual}, maintaining a dataset of points from the current time step and randomly sub-sampled previous points.



     
     

\begin{figure}[h]
\centering

\begin{tabular}{cc}
\includegraphics[width=.38\linewidth]{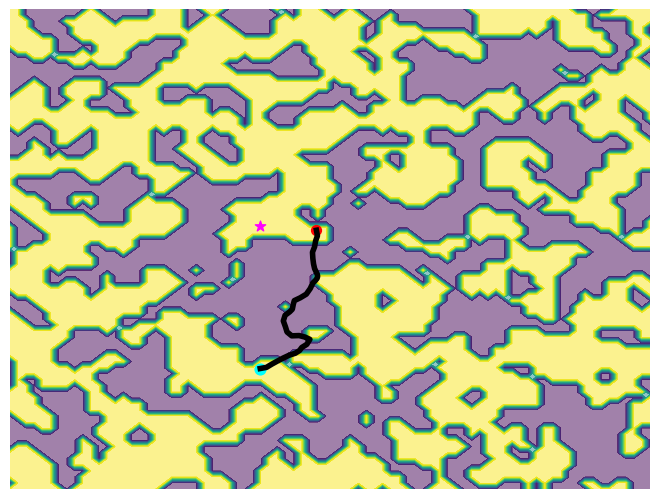}
&
\includegraphics[width=.38\linewidth]{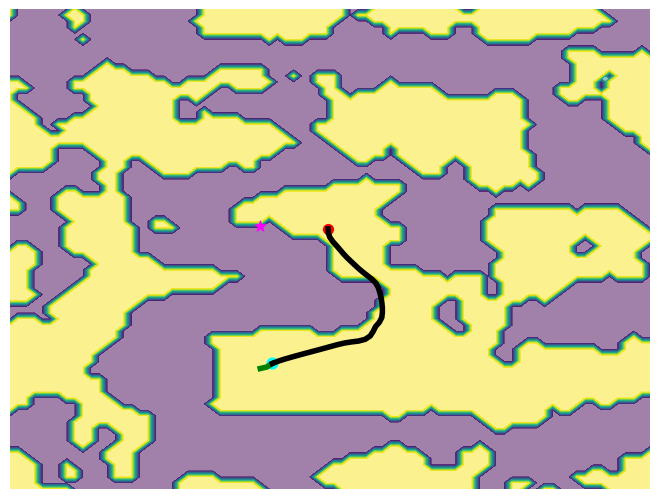}\\

\includegraphics[width=.38\linewidth]{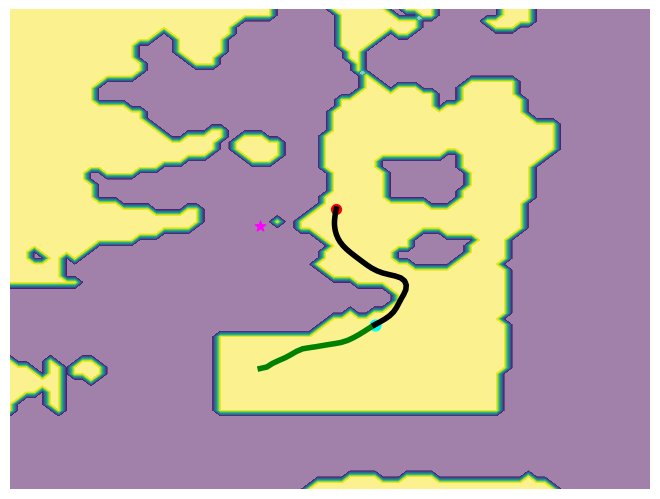} 
&

\includegraphics[width=.38\linewidth]{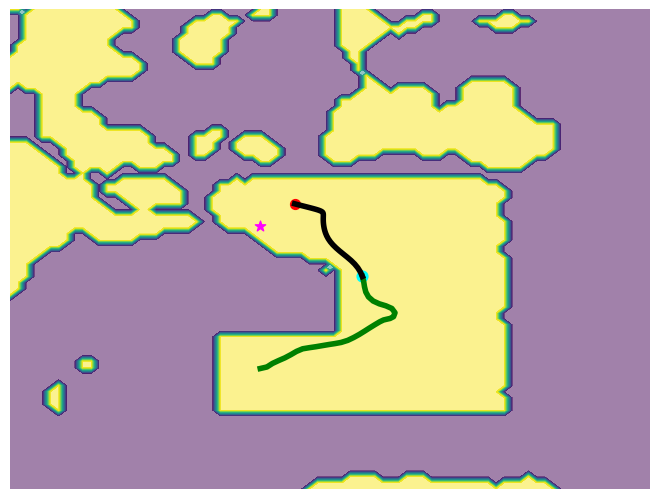}\\

\end{tabular}
\caption{\small Learned map as time progresses and the exploration path chosen at each time step. Green indicates the path taken by the robot, the cyan dot is the robots current position, the red dot is its planned goal, which changes as the map evolves.}
\label{fig:trajectory_explore}
\end{figure}

We ran our setup in a house environment and a factory environment, as shown in Fig~\ref{fig:final_meshes}. For the house, the robot began in the far room in the top right corner and then proceeded to explore each room before ending in the main hallway. Similarly, for the factory, the robot began in the bottom right corner and then looped around each obstacle before terminating in the bottom left corner. The resulting global maps are shown in Fig~\ref{fig:final_meshes} and were generated using 124 local maps and 76 local maps for the house and factory, respectively. Each map took on average 3.5 minutes with a maximum training time of 29 minutes and a minimum of 5 seconds. We believe that the maximum training time corresponded to the robot suddenly entering a large room, while the minimum training time was most likely due to a mostly trained network fluctuating around the caching threshold. Note that even in the 29 minute case, while the network was training the robot was still navigating and ingesting new LiDAR scans.  The average distance traveled between incorporating a new scan to the dataset was $0.168$ meters for the house and $0.13$ meters for the factory. 

As expected, the batched baseline reached its early cutoff condition before it could converge to a good map of the environment and the fine-tuned network experienced catastrophic forgetting, resulting in it only being able to learn a representation for the last location visited by the robot. Surprisingly, Continual Neural Rendering \cite{yan2021continual} is unable to learn an SDF of the environment. We believe that this is because, unlike the results in \cite{yan2021continual} we are not training with surface normal information, and we are training with LiDAR rather than RGB-D data. 

Lastly, we integrated a receding  horizon  model  predictive  controller with our approach and had a robot dynamically determine the trajectory needed to reach a known goal state, given its current position, as shown in Fig.~\ref{fig:trajectory_explore}. This experiment was run through the Webots simulator. We designed the control input constraint set $\mathbb{U}$ so that the robot moved slowly enough to give our approach time to learn the map of the environment. These commands were then sent to the robot in WeBots every 5 seconds and were recalculated after each command window. As illustrated by Fig.~\ref{fig:trajectory_explore}, our approach is able to automatically update the map online while allowing the robot to at the same time re-plan its trajectory based on the new information. Furthermore, we find that our method naturally encourages the robot to explore the environment because it has to maneuver through circuitous routes to avoid collisions with spurious artifacts in a partially learned SDF map.  As the map quality improves, the trajectories become more direct, and the robot naturally explores less. 


\section{Conclusions}
\label{Sec:Conclusions}
We demonstrate an approach to autonomously learn and navigate in an unknown environment by learning a global SDF map, and employing a fast nonlinear trajectory optimizer.  The global learned SDF is composed of a collection of confidence weighted local learned SDF networks.  We learn the local maps online from LiDAR point cloud data, without surface normal information. We effectively plan and control the robot online using the learned global map by navigating towards unexplored regions. We compare against baseline approaches from recent literature and find that our approach both yields higher fidelity maps and lends itself to efficient online model predictive control trajectory optimization.

In future work we would like to simultaneously estimate the robot's pose in the map, giving a full deep SDF SLAM system. We would also like to explore hardware and software optimization to further improve the update rate of our algorithms to allow for applications to robots with faster dynamics, like a quadrotor or the freeflyer on the International Space Station.

%

\newpage \vfill \eject
\newpage

{
\bibliographystyle{IEEEtran}
\bibliography{ref}
}

\end{document}